\documentclass[review]{elsarticle}
\usepackage{graphicx}

\usepackage{tikz}
\usepackage{comment}
\usepackage{booktabs}
\usepackage{multirow}
\usepackage{amsmath,amssymb} 
\usepackage[colorlinks,
            linkcolor=red,      
            anchorcolor=green,  
            citecolor=blue]{hyperref}

\journal{Journal of \LaTeX\ Templates}

\begin{document}

\begin{frontmatter}

\title{AFE-CNN: 3D Skeleton-based Action Recognition with Action Feature Enhancement}

\author{Shannan GUAN, Haiyan LU, Linchao ZHU, Gengfa FANG}
\address{Australia Artificial Intelligence Institute}
\address{School of Electrical and Data Engineering}
\address{University of Technology Sydney, Australia, AU}




\begin{abstract}
Existing 3D skeleton-based action recognition approaches reach impressive performance by encoding handcrafted action features to image format and decoding by CNNs.
However, such methods are limited in two ways: 
a) the handcrafted action features are difficult to handle challenging actions, and b) they generally require complex CNN models to improve action recognition accuracy, which usually occur heavy computational burden. 
To overcome these limitations, we introduce a novel \emph{AFE-CNN}, which devotes to enhance the features of 3D skeleton-based actions to adapt to challenging actions.
We propose feature enhance modules from key joint, bone vector, key frame and temporal perspectives, thus the \emph{AFE-CNN} is more robust to camera views and body sizes variation, and significantly improve the recognition accuracy on challenging actions.
Moreover, our \emph{AFE-CNN} adopts a light-weight CNN model to decode images with action feature enhanced, which ensures a much lower computational burden than the state-of-the-art methods.  
We evaluate the \emph{AFE-CNN} on three benchmark skeleton-based action datasets: NTU RGB+D, NTU RGB+D 120, and UTKinect-Action3D, with extensive experimental results demonstrate our outstanding performance of \emph{AFE-CNN}.

\end{abstract}

\begin{keyword}
3D Skeleton, Action Recognition, Feature Enhance, Attention
\end{keyword}

\end{frontmatter}

\section{Introduction}

Human action recognition task can be roughly processed as two steps: 1) action features extraction and 2) action classification, wherein the quality of action features largely impact the action classification accuracy.
Early works~\cite{twostream,twostream2,featuredescriptors,salientregion,SpatiotemporalFeature} extract action features from RGB videos, which however suffer from background noise and illumination changes.
Recently, depth camera (e.g., Kinect sensor) emerges as a powerful tool to acquire 3D skeleton data.
Compared with RGB videos, 3D skeleton data are more robust to background noise and illumination changes, and can effectively avoid cluttered backgrounds and irrelevant objects. 
As a result, significant efforts have been devoted in recent year to the 3D skeleton-based human action recognition task.

Existing works using 3D skeleton data apply conventional classifiers, e.g., k-Nearest Neighbor~\cite{knn}, Support Vector Machine~\cite{svm}, which can easily recognize actions by feeding raw skeletons. 
However, these conventional methods~\cite{PAM+PTF,LieGroup} cannot adapt to challenging actions such as human-to-human interaction, human-to-object interaction and are infeasible to apply on large scale datasets~\cite{PoT2I+Inception-v3}. 
To overcome these short comes, recent efforts~\cite{tsrji,MLSTM+WeightFusion,GFT,ST-LSTM,Multi-TaskCNNwithRotClips} adopt deep learning-based methods (e.g., recurrent neural network (RNN)~\cite{Logsig-RNN}, long short term memory (LSTM)~\cite{ST-LSTM}, graph convolutional network (GCN)~\cite{stgcn} and convolutional neural network (CNN)~\cite{tsrji}) to recognize actions.
Amongst deep models, RNN, LSTM, and GCN-based models naturally suffer from increasing model complexity (e.g., a large number of input features) and computational burden to improve action recognition accuracy. 
Moreover, they are easily trapped in overfitting problems and cannot directly learn high-level features with spatio-temporal information~\cite{PoT2I+Inception-v3}. 

On the contrary, CNN-based methods encode high-level features and spatio-temporal information in an image, which can effectively reduce model overfitting and represent 3D skeleton data in a more comprehensive way. 
The core of developing CNN-based action recognition models is to effectively encode features from 3D skeleton data so as to improve the recognition accuracy~\cite{3scaleResNet152}. 
Existing approaches to CNN-based action recognition adopt handcrafting approaches~\cite{TCN,ImageGeneration+VGG-19,PoT2I+Inception-v3,SPMFInception-ResNet-222,skeletonmotion,SynthesizedCNN,3scaleResNet152} to encode features to images. 
Specifically, these handcrafting approaches focus on compressing more features (e.g., skeleton motion~\cite{skeletonmotion}, joint angles~\cite{PoT2I+Inception-v3}) in one image and employee a state-of-the-art CNN architecture to decode features and recognize actions. 
However, such handcrafted images are difficult to handle challenging data which consist various of camera views, body sizes and marginally different actions (e.g., writing and typing).

In this paper, we propose the \emph{AFE-CNN}: a learning-based action feature enhance model to enhance features of 3D skeleton-based actions for better encoding of actions to image format.  
Specifically, compared with handcrafted action feature-based images, our AFE-CNN utilizes learning-based methods to enhance features of 3D skeleton-based actions from key joints and bone vectors perspectives, which makes the action recognition model more robust to various camera views and body sizes.
Then, a light-weight four-stream CNN model is deployed to learn comprehensive information in action feature-enhanced images and recognize actions. 
Notably, our designed action feature enhance modules can effectively improve model generalization on challenging actions (e.g., reading, writing). 
Furthermore, our AFE-CNN emphasizes key frames in skeleton sequences by a frame attention module, and we also embed temporal information in transformed images for enhancing the temporal information.
Finally, we train our AFE-CNN end to end and achieves 86.2\% on cross-subject metric and 92.2\% on cross-view metric on the benchmark dataset NTU-RGB+D. 
Furthermore, our AFE-CNN not only achieves outstanding performance, but also very low computational costs, e.g., it costs 3.5ms for one forward inference. Our main contributions are summarized as three folds:
\begin{itemize}
    \item We propose a novel learning-based model to enhance the features of 3D skeleton-based actions, which achieves the state-of-the-art performance on three benchmark datasets for the action recognition task. 
    \item We design action feature enhance modules to adapt to various camera views and body sizes, thus significantly improves the recognition accuracy of our proposed model on challenging actions.
    \item Our AFE-CNN adopts a light-weight CNN architecture to alleviate the computation burden and can largely reduce computing time.
\end{itemize}

\section{Related Works}
The most relevant works to our studies can be roughly classified into two categories.
First category contains deep learning-based approaches for 3D skeleton-based action recognition, and the second category contains the action features via visual representation approaches for 3D skeleton-based action recognition.
\subsection{Deep Learning-based Action Recognition}
To recognize an action, it generally requires spatio-temporal information in a 3D skeleton sequence. 
Therefore, RNN and LSTM network can demonstrate their advantages on spatio-temporal information processing. 
Liao et al.~\cite{Logsig-RNN} encoded the spatio-temporal information from streamed 3D skeleton data, wherein a novel hybrid RNN architecture decode them to recognize actions. 
To reduce the effect on irrelevant joints, Liu et al.~\cite{GCALSTM} designed a global context-aware attention LSTM to selectively learn the spatio-temporal information from most relevant joints. 
In~\cite{SR-TSL}, Si et al. novelty combined a spatial reasoning network and a temporal stack learning network to learn the spatial structural information and temporal information of skeleton sequences. 
With the similar idea of combining networks, Zhang et al.~\cite{TS-LSTM} combined two streams LSTM networks, one is to decode spatio-temporal information in skeleton sequence, another is to transform viewpoints to address various of camera views.

From the perspective of 3D skeleton structure, the key joints are connected with each other by bone links, which contain rich spatial information. 
Hence, the GCN-based model shows another way of thinking to represent spatio-temporal information in a 3D skeleton sequence.
Li et al.~\cite{AS-GCN} adopted a "two stream" strategy and proposed an action-structural graph convolution network to capture actional links and structural links, which can learn both spatial and temporal features for action recognition. 
To capture variation of spatio-temporal information in 3D skeleton data, Gao et al.~\cite{GR-GCN} utilized spatio-temporal modeling of 3D skeleton data and applied optimization on consecutive frames for efficient spatio-temporal data representation.
In~\cite{GVFE+DH-TCN}, Papadopoulos et al. adopted two novel GCN-based models to capture vertex features and short/long-term temporal features for action recognition.
However, such GCN-based methods generally have a low computational efficiency due to their complex network structures.

\subsection{Action Features via Visual Representation}
Generally, a 3D skeleton sequence can be easily encoded into one image and further decoded by a CNN-based model, thus the recent CNN-based works mostly make effort on exploring features in 3D skeleton data and represent them by images. 
Kim et al.~\cite{TCN} directly transformed frame-wise 3D coordinates of joints to images and decoded them by using a residual temporal CNN model.
In~\cite{ImageGeneration+VGG-19}, Li et al. manually transform raw 3D skeleton data to images
by utilizing joint reference and projection, and employed a pre-trained VGG-19 model for end to end training.
Despite the success on utilizing high level features, representing low-level features (e.g., joint angles, motion direction and magnitude) are also considered as an effective approach to improve the accuracy of 3D skeleton-based action recognition.
Kim et al.~\cite{PoT2I+Inception-v3} represented the features by using the spatial correlations of joints and temporal dynamics of 3D skeleton, and achieved remarkable performance. 
To eliminate the effect of camera view variations, Liu et al.~\cite{SynthesizedCNN} presented an skeleton feature enhanced method for view invariant action recognition, to further improve the performance, the skeleton motion enhancement methods are also applied.
In~\cite{3scaleResNet152}, Li et al. separated the skeleton into several body parts and mapped them to images via a scale invariant transformation approach to eliminate the effect of different body sizes.
Combining GCN and CNN is also a novel approach to explore spatio-temporal information in 3D skeleton data.
Zhang et al.~\cite{SGCN} developed an simple yet effective GCN-based model to enhance the joint feature by introducing semantics of joints and hierarchically exploit their relationship, and used CNN model to explore correlation across frames.

Generally speaking, the spatio-temporal information in 3D skeleton data can be easily encoded in image format and effectively decoded by CNN-based models. 
However, the existing approaches using handcrafted action feature images are difficult to handle challenging actions and generally require a large scale CNN model to decode spatio-temporal information in image patterns. 
Therefore, it is highly desirable to develop a more efficient method to encode action features that can be decoded by a light-weight CNN model for reducing computational burden.

\section{Methodology}

\begin{figure}[h]
\centering
\includegraphics[height=4cm ,width=12.1cm,angle=0]{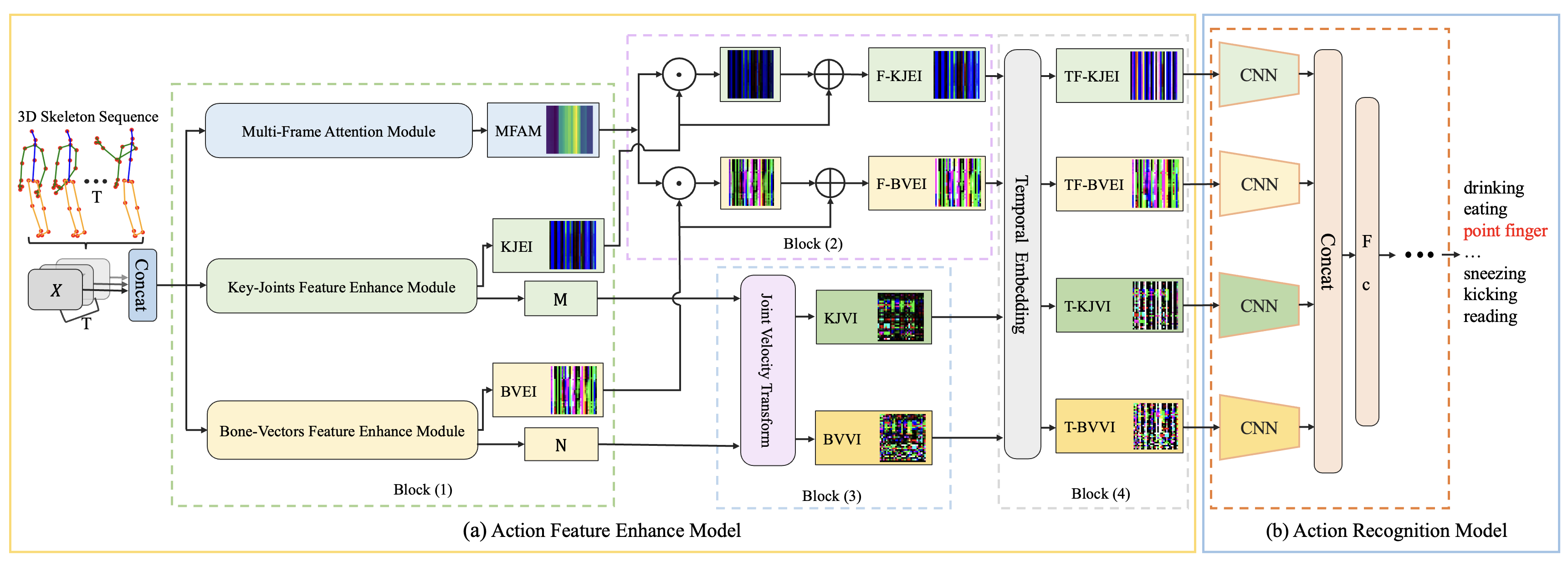}
\caption{Pipeline of our AFE-CNN. First, the 3D skeleton sequence is initially transformed into four images by the Action Feature Enhance module. Then, these four images are fed into a CNN-based Action Recognition Model for action recognition.}
\label{fig:generalpipline}
\end{figure}

\subsection{Model Overview}

As shown in Fig~\ref{fig:generalpipline}, our AFE-CNN contains two main models: Action Feature Enhance model and Action Recognition model. 
In the Action Feature Enhance model, there are four blocks: 1) Skeleton Sequence Image Transform block; 2) Multi-Frame enhance block; 3) Skeleton Motion Velocity Image Transform block and 4) Temporal Embedding block.

In Block 1, the input is 3D skeleton sequences and the outputs contain two parts. The first part has three images, namely Multi-Frame Attention Map (MFAM), Key-Joints Feature Enhance Image (KJEI) and Bone-Vectors Feature Enhance Image (BVEI). 
Specifically, MFAM is calculated from skeleton frames through the Multi-Frame Attention Module. 
KJEI specifies key joints of 3D skeleton sequences
which is produced by the Key-Joints Feature Enhance Module.
BVEI specifies bone vectors of 3D skeleton sequences through a Bone-Vectors Feature Enhance Module.
The second part contains two matrices, namely $\mathbf{M}$ and $\mathbf{N}$, where $\mathbf{M}$ represents the enhanced key joints by the Key-Joints Feature Enhance Module, and $\mathbf{N}$ represents the enhanced bone vectors from the skeleton sequence by the Bone-Vectors Feature Enhance Module.
In Block 2, the inputs are MFAM, KJEI and BVEI from block 1 and the outputs are two new images, namely the frame-enhanced KJEI (F-KJEI) and the frame-enhanced BVEI (F-BVEI). 
In Block 3, the inputs are $\mathbf{M}$ and $\mathbf{N}$. The outputs are two images, namely Key-Joints Motion Velocity Image (KJVI) and Bone-Vectors Motion Velocity Image (BVVI), which are generated by the joint velocity transform model.
In Block 4, the Temporal Embedding module takes four images from blocks 2 and 3 as the inputs and generates four feature enhanced action-transformed images, namely temporal frame-enhanced KJEI (TF-KJEI), temporal frame-enhanced BVEI (TF-BVEI), temporal-enhanced KJVI (T-KJVI), and temporal-enhanced BVVI (T-BVVI), which include the temporal information of the skeleton sequence.
The action recognition model consists of four light-weight convolutional neural networks and several fully connected layers. 
The inputs are the four enhanced action-transformed images, i.e., TF-KJEI, TF-BVEI, T-KJVI, and T-BVVI from Block 4 and the output is the action label.

\begin{figure}[h]
\centering
\includegraphics[height=6cm ,width=12cm,angle=0]{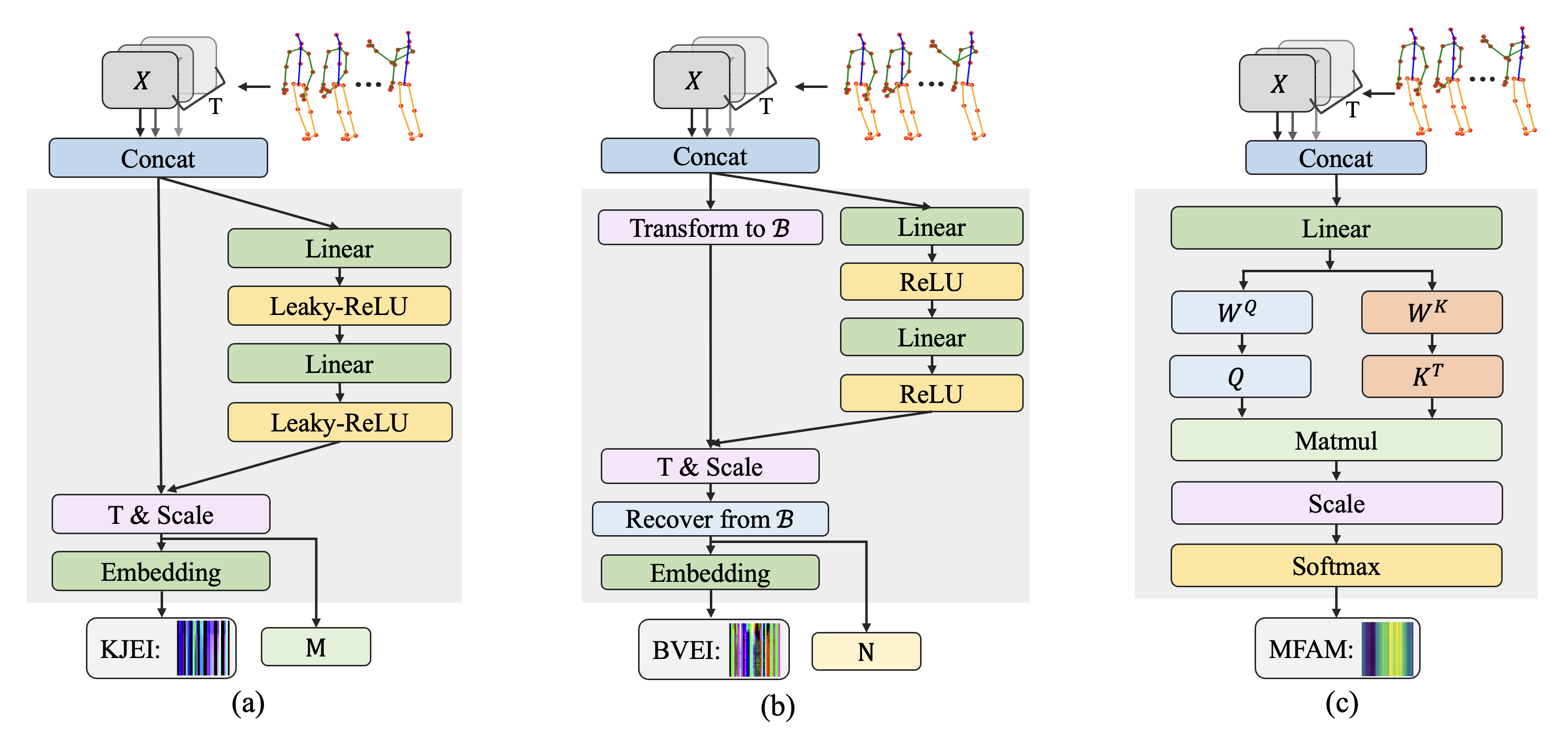}
\caption{Detailed structures of proposed three modules, (a) Key-Joint Feature Enhance Module, (b) Bone-Vectors Feature Enhance Module, and (c) Multi-Frame Attention Module.}
\label{fig:modules}
\end{figure}

\subsection{Action Feature Enhance Model}
In this section, we will discuss the details about five modules in the action feature enhance model.
\subsubsection{Key-Joints Feature Enhance Module} 
Generally, actions with subtle limb motion differences could only cause pixel-level differences among their action-transformed images. 
For instance, the transformed image of ``writing'' is slightly different from the counterpart of ``typing'', since the limb motion differences only reflect by their fingers.  
As a result, 
such subtle motion differences could not be picked up by CNN-based models, leading a sub-optimal performance in action recognition. 
To enlarge the differences among the transformed images of actions, we propose a Key-Joints Feature Enhance Module that uses the key-joints of given skeleton sequences to enhance features of transformed images. 

Let $\mathcal{X}=\left\{\mathbf{X}^{t} \in \mathbb{R}^{3 \times J} \mid t=1, \ldots, T\right\}$ denote the skeleton frames in a given skeleton sequence, 
where $\mathbf{X}^{t}$ is a matrix with size $3 \times J$ that represents 3D coordinates of $J$ joints in frame $t$, and $T$ is the number of frames in the skeleton sequence.
As shown in Fig~\ref{fig:modules} (a), the $\mathcal{X}$ firstly passes a concatenation step to form a $T \times J \times 3$ tensor. 
This tensor then passes two fully connected layers with leakyReLU~\cite{leakyrelu} activation function and outputs a scale matrix $\mathbf{W} \in \mathbb{R}^{1\times J \times 1}$. 
In the parallel, the tensor is transformed to a matrix of size $J \times T \times 3$, which is treated as a 3-channel image of size $J \times T$ sized 3-channel image: $\widetilde{\mathbf{X}} \in \mathbb{R}^{3 \times J \times T}$. 
Finally, the scaled joints matrix is calculated by:
\begin{equation}
\mathbf{M} = \mathbf{W} \times \widetilde{\mathbf{X}}
\end{equation}

Thereafter, the $\mathbf{M}$ is fed to an Embedding layer for linear transformation, where the embedding layer is a $T \times J$ weight matrix, to produce the key-joints feature enhanced image  $\widetilde{\mathbf{M}} \in \mathbb{R}^{3 \times T \times T}$, which is referred to as KJEI. 

\subsubsection{Bone-Vectors Feature Enhance Module} 

Having captured key-joints features through our Key-Joints Feature Enhance Module, we further include a Bone-Vectors Feature Enhance Module to enhance the transformed images using bone vector features. 
The bone vectors contain more details of human actions that could not be captured by less ideal camera views, e.g., the ones taken by cameras from the back-side. 
A camera standing behind humans can only film the human back, and information about human hands motion and gestures cannot be captured. 
Fortunately, the bone vectors can provide integral human body maps to fill the losing motion information. 
We now introduce our Bone-Vectors Feature Enhance Module that includes bone vectors of humans to enhance the transformed images of actions. 

Let $\boldsymbol{p}_{k}$ denotes the 3D coordinate of $k$-th joint and $\boldsymbol{b}_{k}$ denotes the bone vector formed by the $r$-th joint and $q$-th joint, which can also be represented as $\mathbf{X}\boldsymbol{c}$, and calculated as follows:
\begin{equation}
\boldsymbol{b}_{k}=\boldsymbol{p}_{r}-\boldsymbol{p}_{q}=\mathbf{X}\boldsymbol{c}_{k}
\end{equation}
where $\boldsymbol{c}_{k}=(0, \ldots, 0,1,0, \ldots, 0,-1,0, \ldots, 0)^{T}$, with $1$ and $-1$ indicating $\boldsymbol{p}_{r}$ and $\boldsymbol{p}_{t}$, respectively. 
For each $\mathbf{X}^{t}$, bone vector $\boldsymbol{b}_{k}^{t}$ is calculated and concatenated to form $\mathbf{B}^{t}=\left(\boldsymbol{b}_{1}^{t}, \boldsymbol{b}_{2}^{t}, \ldots, \boldsymbol{b}_{b}^{t}\right) \in \mathbb{R}^{3 \times b}$ based on the formula below:
\begin{equation}
\mathbf{B}^{t} = \mathbf{X}^{t}\mathbf{C}
\end{equation}
where $t \in \{1, \ldots, T\}$ is the $t$-th frame in the skeleton sequence; 
$\mathbf{C} \in \mathbb{R}^{J \times b}$ is obtained by concatenating the corresponding vectors $\{\boldsymbol{c}_{1}, \cdots, \boldsymbol{c}_{J}\}$. 
As such, the bone vector matrices of $T$ frames is defined as $\mathcal{B}=\left\{\mathbf{B}^{t} \in \mathbb{R}^{3 \times b} \mid t=1, \ldots, T\right\}$. 

As shown in Fig~\ref{fig:modules} (b), the $\mathcal{X}$ passes similar steps to the ones in Fig~\ref{fig:modules} (a) except for the ``transform to B'' step which outputs a matrix $\mathbf{V} \in \mathbb{R}^{1\times b \times 1}$ for scaling scales bone vector lengths. 
This scaled bone vector collection $\mathcal{B}$ is converted into a matrix of size $b \times T$ : $\widetilde{\mathbf{B}} \in \mathbb{R}^{3 \times b \times T}$ with 3 channels, and the scaled bone matrix is calculated by:
\begin{equation}
\mathbf{N}_{v} = \mathbf{V} \times \widetilde{\mathbf{B}}
\end{equation}

After that, the scaled bone matrix is manipulated to recover the key joints position by:
\begin{equation}
\mathbf{N} = \mathbf{C}^{-1} \mathbf{N}_{v} + \boldsymbol{p}_{0}
\end{equation}
where $\boldsymbol{p}_{0}$ is the root joint position. 
Finally, we fed $\mathbf{N}$ to the embedding layer which is a $T \times J$ weight matrix to produce the bone-vectors feature enhanced image $\widetilde{\mathbf{N}} \in \mathbb{R}^{3 \times T \times T}$, which is also referred to as BVEI.

\subsubsection{Multi-Frame Attention Module (MFAM)}

Generally, the label of an action depends on a series of key skeleton frames, which means every single frame in an action sequence should have different weights in classifying an action. 
For example, giving an action sequence that denotes the action of ``drinking'', it is essentially the ``drink'' frames that decide the action's label instead of ``picking up the cup''.  
Inspired by self-attention mechanism in Transformer module~\cite{attentionisallyouneed}, we specify different importance scores of skeleton frames to emphasize key frames for CNN-based action classification.
We now introduce our Multi-Frame Attention Module that devices the self-attention mechanism to yield different weights for skeleton-transformed images.

From Fig~\ref{fig:modules} (c), the $\mathcal{X}$ is firstly fed to a fully connected layer, then it splits to two branches: 
the left branch $W^{Q}$ is used for providing query matrix $\mathbf{Q} \in \mathbb{R}^{T \times J}$ and the right branch $W^{K}$ for providing key matrix $\mathbf{K} \in \mathbb{R}^{T \times J}$. 
We compute the dot products of $\mathbf{Q}$ and transpose of key matrix $\mathbf{K}^{T} \in \mathbb{R}^{J \times T}$, then divide the dot products by $\sqrt{\boldsymbol{d}_{k}}$, where $\boldsymbol{d}_{k}$ denotes the dimension of key matrix. 
Formally, 
\begin{equation}
\mathbf{A}=\operatorname{softmax}\left(\frac{\mathbf{Q} \mathbf{K}^{T}}{\sqrt{\boldsymbol{d}_{k}}}\right)
\end{equation}
where $\mathbf{A} \in \mathbb{R}^{T \times T}$ is the output multi-frame attention map.

Thereafter, as shown in Fig~\ref{fig:generalpipline}, we multiply the multi-frame attention map $\mathbf{A}$ with key-joints feature enhanced image $\widetilde{\mathbf{M}}$. 
To prevent loss information, we add $\widetilde{\mathbf{M}}$ itself after multiplication, and the final key-joint attention feature image is computed by:
\begin{equation}
\operatorname{ Attention }(\mathbf{Q}, \mathbf{K}, \widetilde{\mathbf{M}})= \widetilde{\mathbf{M}} \odot \mathbf{A}  + \widetilde{\mathbf{M}}
\end{equation}
where $\odot$ is the element-wise product and $\operatorname{ Attention }$ is the attention operator in our Multi-Frame Attention Module. Similarly, the bone-vector attention feature image is obtained by:
\begin{equation}
\operatorname{ Attention }(\mathbf{Q}, \mathbf{K}, \widetilde{\mathbf{N}})=\widetilde{\mathbf{N}} \odot \mathbf{A} + \widetilde{\mathbf{N}}
\end{equation}

\subsubsection{Joint Velocity Image Transform Module}
Inspired by two-stream architecture~\cite{twostream}, which utilizes optical flow fields to complement the original stream.  
We propose a Skeleton Moving Velocity Transform module to calculate moving velocity of each key joint as another action feature representation stream.
Given a key joint position as $\boldsymbol{p}=(x, y, z)$, a $J$-joints skeleton is denoted as $\mathbf{S}=\left\{\boldsymbol{p}_{1}, \boldsymbol{p}_{2}, \ldots, \boldsymbol{p}_{J}\right\}$. 
The moving velocity of each key joint between two frames is computed by:
\begin{equation}
\mathbf{E}=\frac{\mathbf{S}^{t+1}-\mathbf{S}^{t}}{\Delta T}=\left\{\frac{\boldsymbol{p}_{1}^{t+1}-\boldsymbol{p}_{1}^{t}}{\Delta T}, \frac{\boldsymbol{p}_{2}^{t+1}-\boldsymbol{p}_{2}^{t}}{\Delta T}, \ldots, \frac{\boldsymbol{p}_{J}^{t+1}-\boldsymbol{p}_{J}^{t}}{\Delta T}\right\}
\end{equation}
where $\Delta T$ denotes the time difference between two consecutive frames and $t$ is the frame index. 
As shown in Fig~\ref{fig:generalpipline}, we fed the scaled joints matrix $\mathbf{M}$ and the scaled bone matrix $\mathbf{N}$ into the transform module, which outputs $ J \times T$ image matrices. 
Thereafter, they are transformed to a $3 \times T \times T$ sized image matrix by linear transformation, which named key joint velocity image (KJV) and bone vectors velocity image (BVV), respectively. 

\subsubsection{Temporal Embedding Module(TE)}

It is well-acknowledged that recognizing an action is highly dependent on the timing when the key poses happen~\cite{framhappens}. 
For example, standing and sitting are two timely opposite actions, and one can be regarded as the reversed time sequence of another action. 
A 3D skeleton-based action recognition task generally transforms a skeleton sequence into an action feature image, and the temporal information is represented as the relative position of the pixels and embed in action feature image.  
However, conventional CNN model is considered to be spatial-agnostic~\cite{spatial-info}, which makes it difficult to find the temporal information of an action-transformed image. 
To address this problem, we propose a Temporal Embedding (TE) module to enhance the temporal information in the action-transformed images.

Inspired by ViT~\cite{ViT}, we provide a $1 \times T$ dimensional relative positional embedding as temporal information for action-transformed images. 
Here, we regard each column of an action-transformed image as the corresponding key frame pose, and we utilize the relative distance between columns to encode the spatial information. 
As shown in Fig~\ref{fig:generalpipline}, we add a learned $1 \times T$ dimensional positional embedding to all action-transformed images to enhance the temporal features. 
To this end, these images will be fed to our Four-Stream CNN model. 

\subsection{Action Recognition Model}
In this section, we will discuss the details of CNN models used in our action recognition model and our training strategy.
\subsubsection{Four-Stream CNN}
\begin{figure}[h]
\centering
\includegraphics[height=6cm ,width=12cm,angle=0]{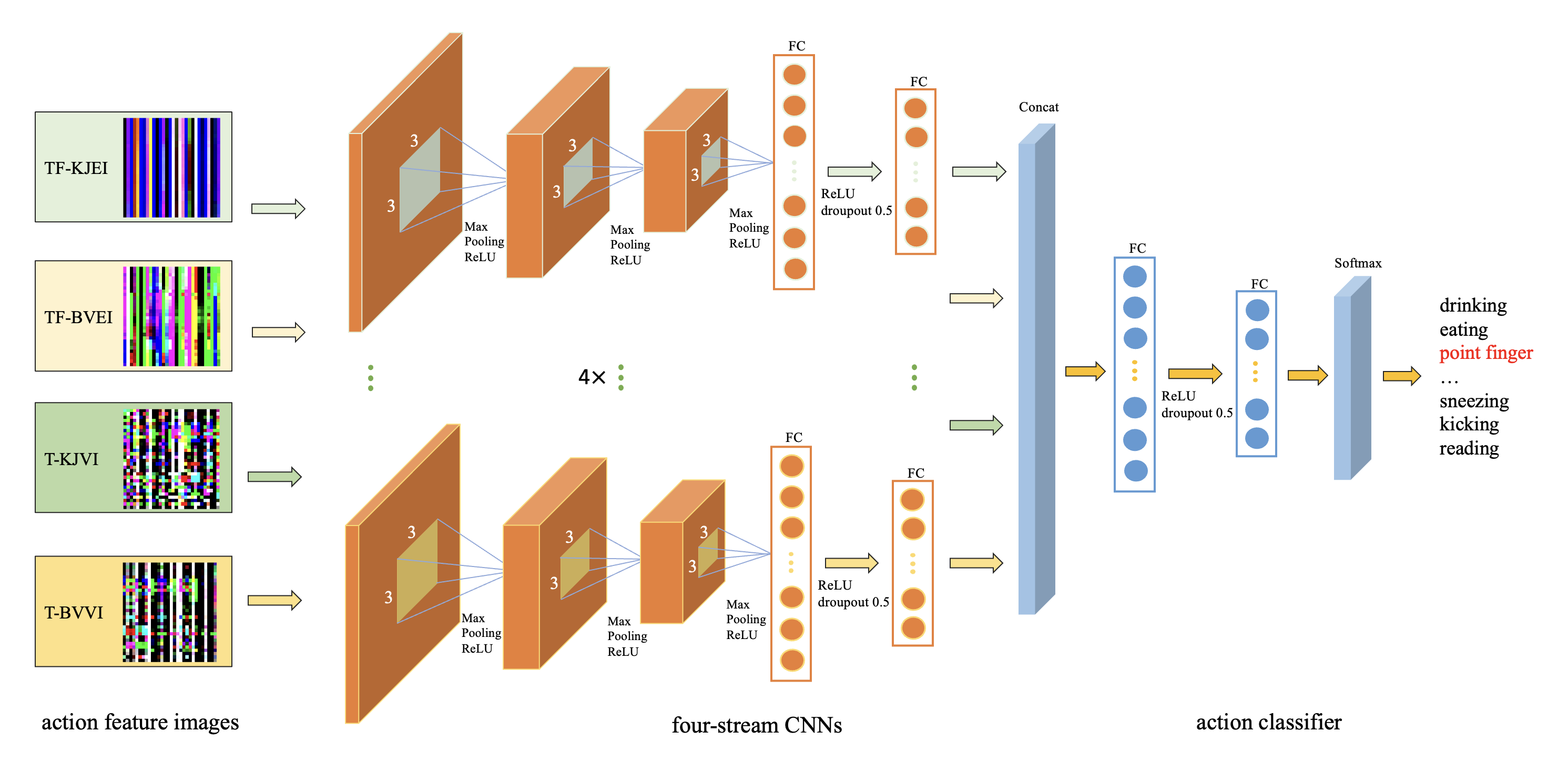}
\caption{Four-stream CNN-based action recognition model.}
\label{fig:cnnmodel}
\end{figure}

To obtain more discriminate features from action-transformed images, we propose a four-stream CNN-based action recognition model. 
Our four-stream CNN model is a modified version of a lightweight CNN model which is proposed by Caetano et al.~\cite{skeletonimagebaseCaetano}. 
This lightweight CNN contains three convolution layers which have similar structure to VGG Net~\cite{VGGnet}, and followed by two fully connected layers.
Here, we modified their convolution layers with different padding size and different dimension of fully connected layers. 
The kernel size for each convolution layer is $3 \times 3$ with a stride of 2. 
After each convolution layer, we adopt a max pooling and a LeakyReLU~\cite{leakyrelu} layer. 
For each action-transformed image, we employ a CNN model to extract the action feature vector, then, we concatenate these four action feature vectors and fed the consolidated one to a two layers fully connected networks for action recognition.

\subsubsection{Model Optimization}
With a skeleton-based action recognition dataset, our CNN-based action recognition architecture is trained uniformly by using categorical cross-entropy loss function:
\begin{equation}
\mathcal{L}_{cce}=-\sum_{i=1}^{c} t_{i} \cdot \log {y}_{i}
\end{equation}
where $c$ is the number of action categories, $t_{i}$ denotes the true value which is either $0$ or $1$, and $y_{i}$ is the Softmax probability for $i_{th}$ action category.

Compared with conventional hand-craft action-transformed image action recognition methods, our proposed methods do not require any manually action feature extraction. 
Moreover, compared with using large-scale CNN model, we do not need any pre-trained models to improve the performance of action recognition. 

\section{Experiments and Discussion}
We implement our experiments in PyTorch framework with single GeForce RTX 3090 GPU. We train our model with Adam optimizer by using the initial learning rate at 0.001, the shrink factor of 0.1 after 20 epochs. 
Due to different size of datasets, we set a batch size of 64 for both NTU RGB+D~\cite{NTURGB60} and NTU RGB+D 120~\cite{NTURGB120}, and a batch size of 8 for UTKinect-Action3D~\cite{UTKinect}. 

\subsection{Datasets}
NTU RGB+D: This dataset is recorded by Kinect v2 sensors and each skeleton is depicted by 3D locations of 25 body joints. 
In details, it contains 56,880 3D skeleton sequences with 40 different human subjects and covers 60 daily action categories, which including single person actions, human-objective interactions, and human-human interactions. 
It is a challenge dataset due to its large scale, diverse action categories and various of camera views. 
We evaluate our method on this dataset by using two official evaluation metrics: cross-subject (subjects with 20 specific IDs are for training and the remaining for testing) and cross-view (samples from camera 1 for testing while the samples from camera 2 and 3 for training).

NTU RGB+D 120: This dataset is the extended version of the NTU RGB+D by adding another 60 challenging action categories. 
Compared with NTU RGB+D, this dataset is more challenge due to more subjects and increased variations of view points. 
In details, it contains totally 114,480 3D skeleton sequences with 106 subjects which performers in a wide range of age distribution. These samples are captured from 155 different camera views and recorded in 32 different scenes.
This dataset provides two evaluation metrics: cross-subject (subjects with 53 specific IDs are for training and remaining for testing) and cross-setup (samples with even collection setup IDs for training while remaining of odd setup IDs for testing).  

UTKinect-Action3D: Compared with NTU RGB+D and NTU RGB+D 120, this dataset is significantly smaller.
It has 200 3D skeleton sequences of 10 daily human-object interactive action categories. Each skeleton is recorded as 20 body joints. 
Due to its small-scale, we utilize the cross-subject (half of the subjects for training and the remaining subjects for testing) as the evaluation metric to prevent potential risk of model overfitting. 

\subsection{Experiment Results}
Here, we compare our AFE-CNN with several SOTA 3D skeleton action recognition methods on NTU RGB+D, NTU RGB+D 120 and UTKinect-Action3D respectively. 

\subsection{Results on NTU RGB+D}

\begin{table}[h]
\small
\centering
\caption{Performance comparisons on NTU-RGB+D }
\resizebox{110mm}{!}{
\begin{tabular}{l|c|ccc|l}
\toprule
\multicolumn{1}{c|}{\multirow{2}{*}{\textbf{Methods}}} &
\multicolumn{1}{c|}{\multirow{2}{*}{\textbf{Architecture}}} &
\multicolumn{2}{c}{\textbf{Accuracy (\%)}} \\ \cline{3-4}

\multicolumn{1}{c|}{} &
\multicolumn{1}{c|}{} &
\multicolumn{1}{c}{\textbf{CrossSubject}} &
\multicolumn{1}{c}{\textbf{CrossView}}  \\ \midrule

PAM+PTF~\cite{PAM+PTF} & PAM & 68.2 & 76.3 \\
TSRJI~\cite{tsrji}  & CNN & 73.3 & 80.3 \\
ImageGen+VGG-19~\cite{ImageGeneration+VGG-19} & CNN & 75.2 & 82.1 \\
ResTCN~\cite{TCN}  & CNN & 74.3 & 83.1 \\
Skelemotion~\cite{skeletonmotion}  & CNN & 76.5 & 84.7 \\
MTLN~\cite{mtln}  & CNN & 79.6 & 84.8 \\
SPMF+ResNet~\cite{SPMFInception-ResNet-222} & CNN & 78.9 & 86.2 \\
Synthesized CNN~\cite{SynthesizedCNN}  & CNN & 80.0	& 87.2 \\
MTCNN+RotClips~\cite{Multi-TaskCNNwithRotClips}  & CNN & 81.1	& 87.4 \\
ST-GCN~\cite{stgcn}  & GCN & 81.5 & 88.3 \\
VA-LSTM~\cite{VA-LSTM}  & LSTM & 80.7 & 88.8 \\
PoT2I+Inception-v3~\cite{PoT2I+Inception-v3}  & CNN & 83.8 & 90.3 \\ 
3scale ResNet152~\cite{3scaleResNet152}  & CNN & 84.6 & 90.9 \\ \midrule

AFE-CNN (Ours)  & CNN & \textbf{86.2} & \textbf{92.2} \\ \bottomrule
\end{tabular}}
\label{nturgb60results}
\end{table}

\begin{figure}[h]
\centering
\includegraphics[height=10cm ,width=12cm,angle=0]{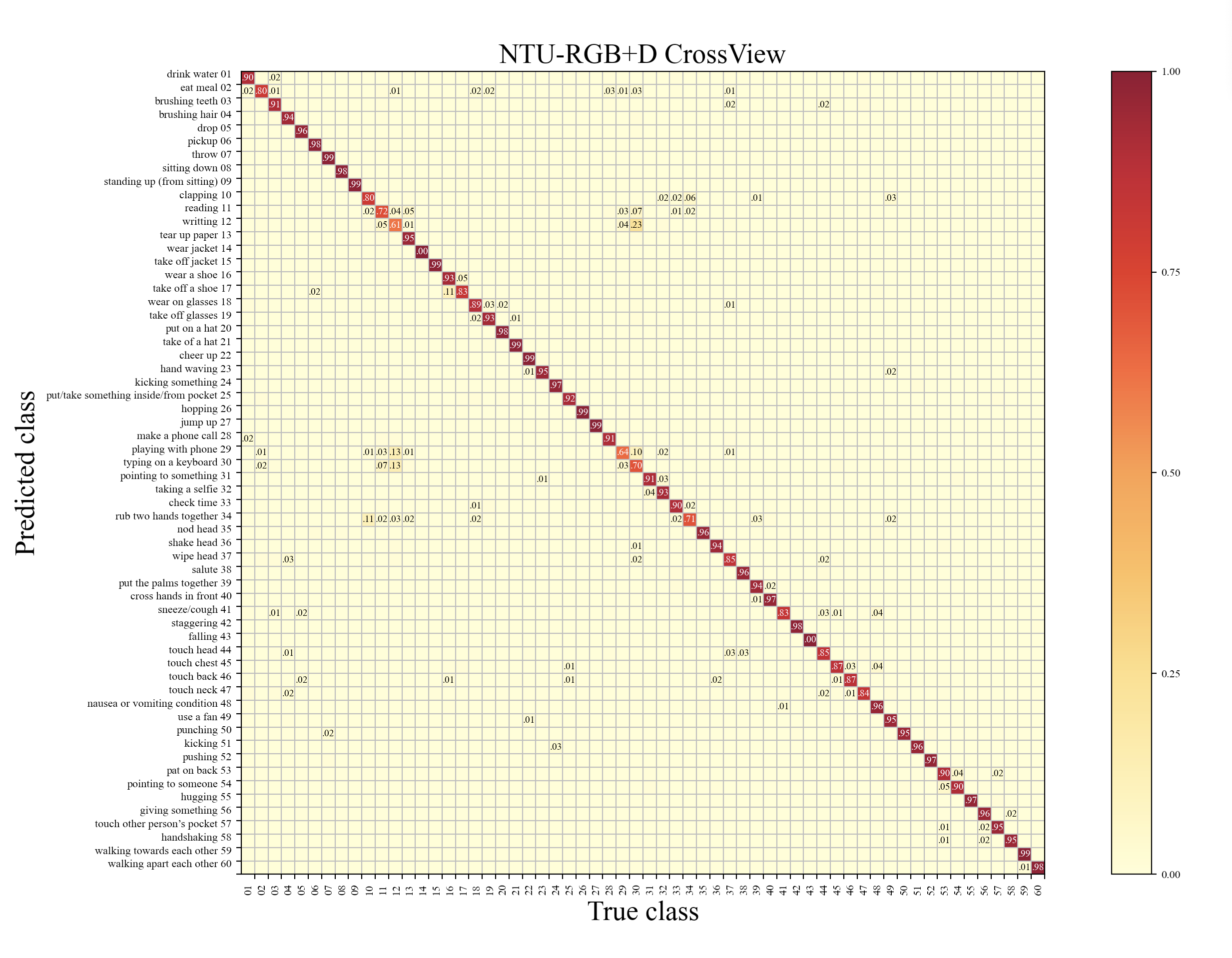}
\caption{The confusion matrix of NTU RGB+D dataset with cross-view evaluation metrics.}
\label{fig:nturgbconfusionmatrixcv}
\end{figure}

\begin{figure}[h]
\centering
\includegraphics[height=10cm ,width=12cm,angle=0]{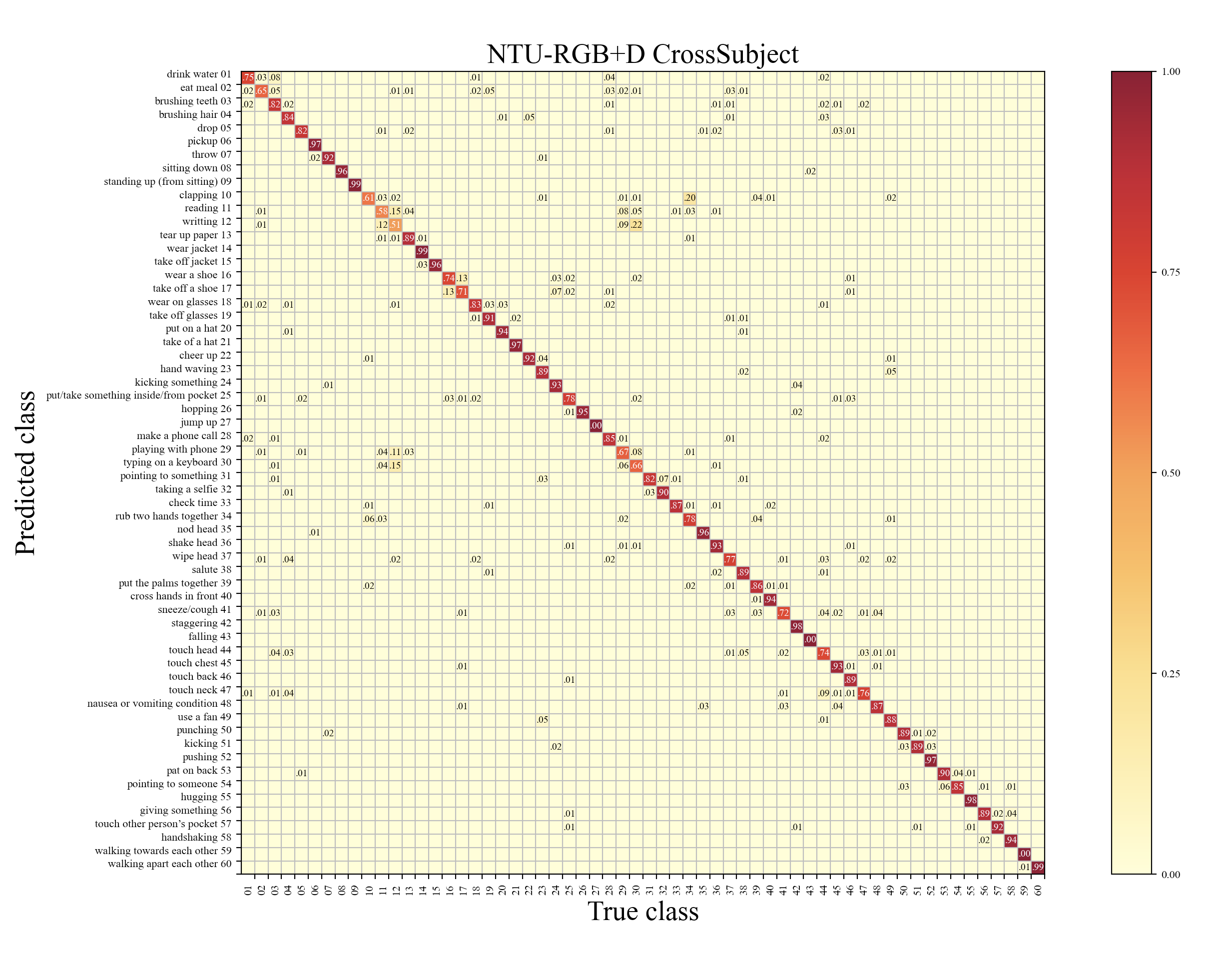}
\caption{The confusion matrix of NTU RGB+D dataset with cross-subject evaluation metrics.}
\label{fig:nturgbconfusionmatrixcs}
\end{figure}

From the results shown in Table~\ref{nturgb60results}, our AFE-CNN achieves the highest accuracy on both cross-subject (86.2\%) and cross-view (92.2\%) evaluation metrics. 
Accordingly, the corresponding confusion matrix of NTU-RGB+D results under cross-subject metric and cross-view metric are depicted in Fig~\ref{fig:nturgbconfusionmatrixcs} and Fig~\ref{fig:nturgbconfusionmatrixcv} respectively.
It can be observed that there is a large gap of accuracy between deep learning-based methods (e.g.,MTLN~\cite{mtln},ResTCN~\cite{TCN}) and traditional machine learning-based methods (e.g., TSRJI~\cite{tsrji} and PAM+PTF~\cite{PAM+PTF}). 
Because the traditional machine learning-based methods are hard to handle large-scale action dataset, therefore, the most of 3D skeleton-based action recognition methods are developed with deep learning-based architectures (e.g., CNN, LSTM, GCN). 

In these deep learning-based approaches, the LSTM-based and GCN-based architecture achieve applaud performance (e.g., ST-GCN~\cite{stgcn}, VA-LSTM~\cite{VA-LSTM}) due to they are expert on spatio-temporal information processing. 
However, compared with CNN-based methods, the LSTM-based and GCN-based methods consume more computing time due to their complex network structure. 

For CNN-based architecture, few of methods directly feed raw coordinate data to a CNN model (e.g., TCN~\cite{TCN}), and the most of methods first transform different levels of action features (e.g., high-level action features~\cite{ImageGeneration+VGG-19,Multi-TaskCNNwithRotClips}, low-level action features~\cite{tsrji,PoT2I+Inception-v3,3scaleResNet152,mtln}, motion features~\cite{skeletonmotion,SPMFInception-ResNet-222,SynthesizedCNN}) to image format as inputs for CNN models. 
To further improve action recognition accuracy, several methods~\cite{3scaleResNet152, PoT2I+Inception-v3,SPMFInception-ResNet-222} employee large scale CNN model (e.g., ResNet, Inception-v3) to extract more spatio-temporal information in large size action feature images. 
Although these approaches significantly improve the performance thanks to the benefits of large scale CNNs feature extraction in image classification, they load more computation burden which largely increase computing time.
Moreover, the hand-crafted action features are difficult to handle challenging data with various of camera views, body sizes, and they have become the bottleneck to further improve the performance.  
In contrast to these methods, our AFE-CNN outperforms all handcrafted action feature image approaches and only requires light-weight CNN models.

\subsection{Results on NTU RGB+D 120}

\begin{table}[h]
\small
\centering
\caption{Performance comparisons on NTU-RGB+D 120}
\resizebox{110mm}{!}{
\begin{tabular}{l|c|ccc|l}
\toprule
\multicolumn{1}{c|}{\multirow{2}{*}{\textbf{Methods}}} &
\multicolumn{1}{c|}{\multirow{2}{*}{\textbf{Architecture}}} &
\multicolumn{2}{c}{\textbf{Accuracy (\%)}} \\ \cline{3-4}

\multicolumn{1}{c|}{} &
\multicolumn{1}{c|}{} &
\multicolumn{1}{c}{\textbf{CrossSubject}} &
\multicolumn{1}{c}{\textbf{CrossSetup}}  \\ \midrule

MTLN~\cite{mtln} & CNN & 58.4 & 57.9 \\
MTCNN+RotClips~\cite{Multi-TaskCNNwithRotClips}  & CNN & 62.2 & 61.8 \\
TSRJI~\cite{tsrji}  & CNN & 67.9 & 62.8 \\
Synthesized CNN~\cite{SynthesizedCNN}  & CNN & 60.3 & 63.2 \\
GCA-LSTM~\cite{GCALSTM} & LSTM & 61.2 & 63.3 \\
Skelemotion~\cite{skeletonmotion}  & CNN & 67.7 & 66.9 \\
Logsig-RNN~\cite{Logsig-RNN} & RNN & 68.3 & 67.2 \\
Gimme Signals~\cite{GimmeSignals}  & CNN & 70.8 & 71.6\\
ST-GCN~\cite{stgcn} & GCN & 70.7 & 73.2 \\
AS-GCN~\cite{AS-GCN} & GCN & 77.7 & 78.9 \\
GVFE+DH-TCN~\cite{GVFE+DH-TCN} & GCN & 78.3 & 79.8 \\
SR-TSL~\cite{SR-TSL} & LSTM & 74.1 & 79.9 \\
SGCN~\cite{SGCN} & GCN & 79.2 & 81.5 \\ \midrule

AFE-CNN (Ours)  & CNN & \textbf{80.4} & \textbf{81.6} \\ \bottomrule
\end{tabular}}
\label{nturgb120results}
\end{table}

As shown in Table~\ref{nturgb120results}, our AFE-CNN outperforms than other methods on both cross-subject (80.2\%) and cross-setup (81.6\%). 
In this dataset, some methods utilize RNN-based\cite{Logsig-RNN} and LSTM-based\cite{GCALSTM,SR-TSL} methods and achieve promising results, and the best LSTM-based~\cite{SR-TSL} method reaches 79.9\% accuracy on cross-setup metric.
In addition, the GCN-based~\cite{AS-GCN,GVFE+DH-TCN,SGCN} action recognition approaches achieve applaud performance due to GCNs are specialize on finding spatio-temporal information in raw 3D skeleton sequence. 
But they still suffer from a high computational burden.
Compared with GCN and LSTM-based methods, our AFE-CNN achieves high accuracy of action recognition while ensuring a low computational burden.  

Moreover, it can be seen that there is a large gap of accuracy between CNN-based methods and GCN-based methods (e.g., Gimme Signals~\cite{GimmeSignals} and GVFE+DH-TCN~\cite{GVFE+DH-TCN}).
This verifies the CNN-based model is hard to handle more challenging dataset due to the limitations of hand-craft features transformed images (e.g., more various of camera views, body sizes and marginally different actions).
However, our AFE-CNN can effectively enhance the action features of challenging actions that enables a light-weight CNN model to achieve outstanding performance.

\subsection{Results on UTKinect-Action3D}

\begin{table}[h]
\small
\centering
\caption{Performance comparisons on UTKinect-Action3D}
\resizebox{100mm}{!}{
\begin{tabular}{l|c|cc|l}
\toprule
\multicolumn{1}{c|}{\textbf{Methods}} &
\multicolumn{1}{c|}{\textbf{Architecture}} &
\multicolumn{1}{c}{\textbf{Accuracy (\%)}} \\  \midrule

MLSTM+Weight Fusion~\cite{MLSTM+WeightFusion} & RNN & 96.0 \\
GFT~\cite{GFT} & GCN & 96.0 \\
ST-LSTM~\cite{ST-LSTM} & LSTM & 97.0 \\
PAM+PTF~\cite{PAM+PTF} & PAM & 97.0 \\
Lie Group~\cite{LieGroup} & SVM & 97.1 \\
PoT2I+Inception-v3~\cite{PoT2I+Inception-v3} & CNN & 98.5 \\
GR-GCN~\cite{GR-GCN} & GCN & 98.5 \\
GCA-LSTM~\cite{GCALSTM} & LSTM & 99.0 \\
MTCNN+RotClips~\cite{Multi-TaskCNNwithRotClips} & CNN & 99.0 \\
Multi-Stream CNN~\cite{Multi-StreamCNN} & CNN & 99.0 \\ \midrule

AFE-CNN (Ours)  & CNN & \textbf{99.0} \\ \bottomrule
\end{tabular}}
\label{UTKinectresults}
\end{table}

\begin{figure}[h]
\centering
\includegraphics[height=6cm ,width=7.2cm,angle=0]{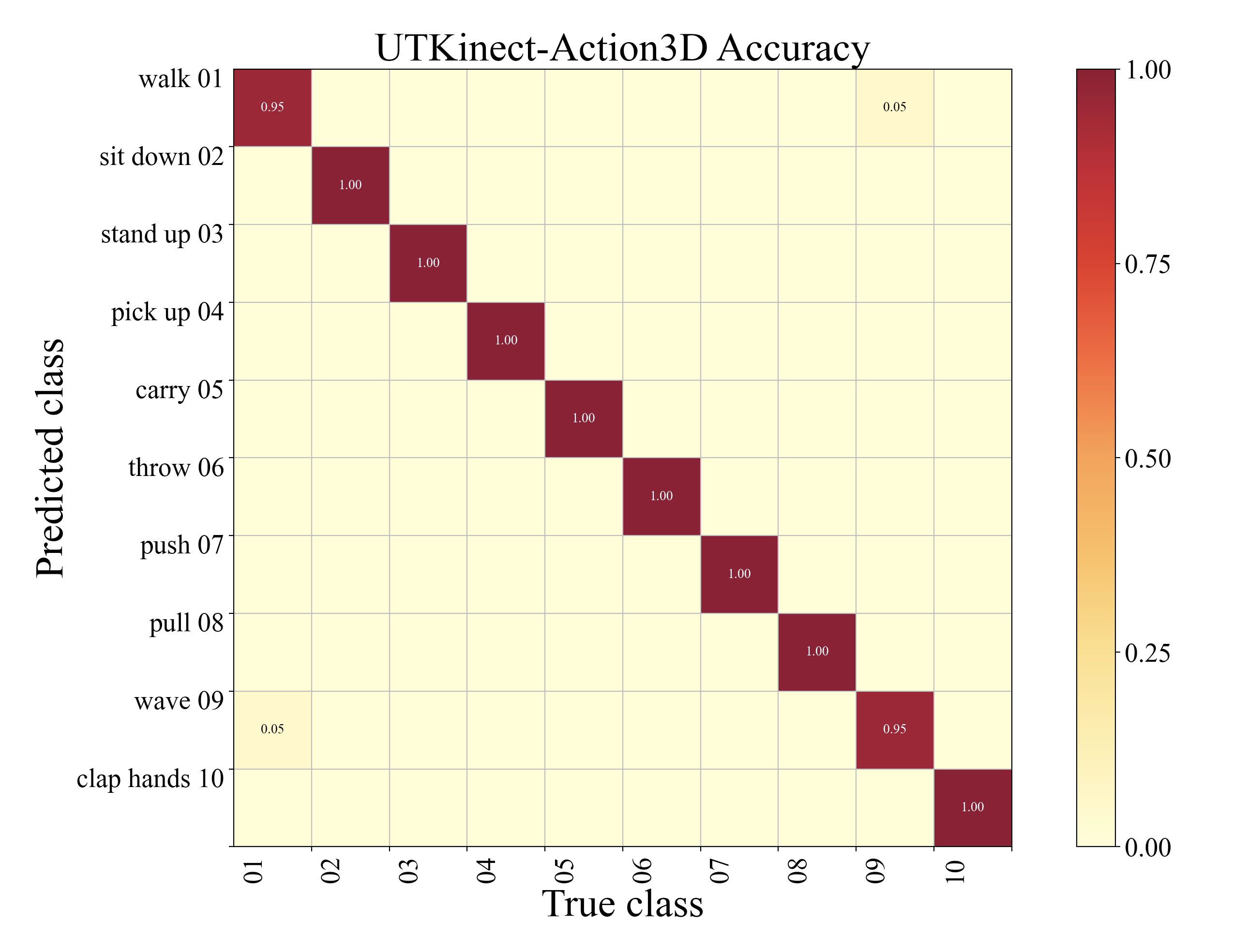}
\caption{The confusion matrix of UTKinect-Action3D dataset.}
\label{fig:utkinectconfusionmatrix}
\end{figure}

From the results shown in Table~\ref{UTKinectresults}, our AFE-CNN achieves the accuracy of 99.0\% in this dataset and several  methods~\cite{GCALSTM,Multi-TaskCNNwithRotClips,Multi-StreamCNN} also achieve the same result. 
Accordingly, the corresponding confusion matrix of this dataset as shown in Fig~\ref{fig:utkinectconfusionmatrix}, where some walk and wave action samples are misclassified to each other. 
In this small-scale dataset, some traditional machine learning-based methods~\cite{PAM+PTF,LieGroup} even achieve comparable performance compared with other deep learning-based methods. 
The CNN-based methods~\cite{PoT2I+Inception-v3,Multi-TaskCNNwithRotClips,Multi-StreamCNN} generally have an applaud performance while the GCN-based~\cite{GFT,GR-GCN} methods have a lower accuracy.  
In these CNN-based methods, only PoT2I+Inception-v3~\cite{PoT2I+Inception-v3} adopts a single CNN architecture, which cannot decode more action features and cause the action recognition accuracy less 0.5\% than other CNN-based methods. 
Although our AFE-CNN utilize a light-weight CNN model, our four-streams CNN architecture ensure that more action features are encoded and decoded, therefore, our AFE-CNN also can achieve a promising performance in small-scale dataset.

\subsection{Ablation Study}
In this section, we carry out several ablation studies on NTU-RGB+D dataset to validate the contribution of key modules of AFE-CNN.

\subsubsection{Ablation study on multi-frame attention and temporal embedding module}
To verify the contributions made by our multi-frame attention module and temporal embedding module, we train the AFE-CNN with and without them and compare the results on the perspective of cross-subject and cross-view metrics.

\begin{table}[h]
\small
\centering
\caption{Ablation study on multi-frame attention and temporal embedding module, AFE-CNN denotes the model without multi-frame attention module (MFAM) and temporal embedding module (TE).}
\resizebox{80mm}{!}{
\begin{tabular}{l|c|cc|l}
\toprule
\multicolumn{1}{c|}{\multirow{2}{*}{\textbf{Methods}}} &
\multicolumn{2}{c}{\textbf{Accuracy (\%)}} \\ \cline{2-3}
\multicolumn{1}{c|}{} &
\multicolumn{1}{c}{\textbf{CrossSubject}} &
\multicolumn{1}{c}{\textbf{CrossView}}  \\ \midrule
AFE-CNN  & 84.9 & 90.1  \\
AFE-CNN+MFAM  & 85.7 & 91.3  \\
AFE-CNN+MFAM+TE   & \textbf{86.2} & \textbf{92.2} \\ \bottomrule
\end{tabular}}
\label{ablationstudyonFAMPE}
\end{table}

As shown in Table~\ref{ablationstudyonFAMPE}, we can observe that the temporal embedding module could improve accuracy on cross-subject and cross-view metrics by 0.5\% and 0.9\% respectively. 
This means the temporal embedding module can effectively enhance the temporal information in action feature images and improves the performance of AFE-CNN.
For multi-frame attention module, it improves the accuracy on cross-subject metric by 0.8\% and the accuracy on cross-view metric by 1.2\%, both of them are higher than temporal embedding module, which means multi-frame attention module contributes more than the temporal embedding module in AFE-CNN.

\subsubsection{Ablation study on Action Feature Enhance Modules}
To verify the contributions of our key-joints feature enhance module and bone-vectors feature enhance module, we train the four-stream CNN-based action recognition model with and without these modules and compare their results on cross-subject and cross-view metrics.

\begin{table}[h]
\small
\centering
\caption{Ablation study of Action Feature enhance modules. AF-CNN only takes raw 3D skeleton transformed images without using any our proposed action feature enhanced module. The KJFE denotes the key joints feature enhance module and BVFE is the bone vectors feature enhance module.}
\resizebox{80mm}{!}{
\begin{tabular}{l|c|cc|l}
\toprule
\multicolumn{1}{c|}{\multirow{2}{*}{\textbf{Methods}}} &
\multicolumn{2}{c}{\textbf{Accuracy (\%)}} \\ \cline{2-3}
\multicolumn{1}{c|}{} &
\multicolumn{1}{c}{\textbf{CrossSubject}} &
\multicolumn{1}{c}{\textbf{CrossView}}  \\ \midrule
AF-CNN & 79.2 & 84.0  \\
AF-CNN+KJFE  & 82.9 & 89.1 \\
AF-CNN+BVFE  & 83.8 & 88.2  \\
AF-CNN+KJFE+BVFE   & \textbf{84.9} & \textbf{90.1} \\ \bottomrule
\end{tabular}}
\label{ablationstudyonKIEBVE}
\end{table}

From the results shown in Table~\ref{ablationstudyonKIEBVE}, we can observe that both of key-joints feature enhance module and bone-vectors feature enhance module can boost the performance of our CNN-based action recognition model.  
However, they show a different improvements on two evaluation metrics. 
The KJFE improves more on cross-view metrics (up to 89.1\%) while BVFE improves more on cross-subject metric (up to 83.8\%). 
This phenomenon is caused by two factors: one is the variations of camera views on two evaluation metrics, and the other is that the BVFE applies weights on bone vectors, which is more effective than key joints on some rare camera views. When combining KJFE and BVFE, we see significant performance improvements on both evaluation metrics.

\begin{figure}[h]
\centering
\includegraphics[height=3.8cm ,width=12.1cm,angle=0]{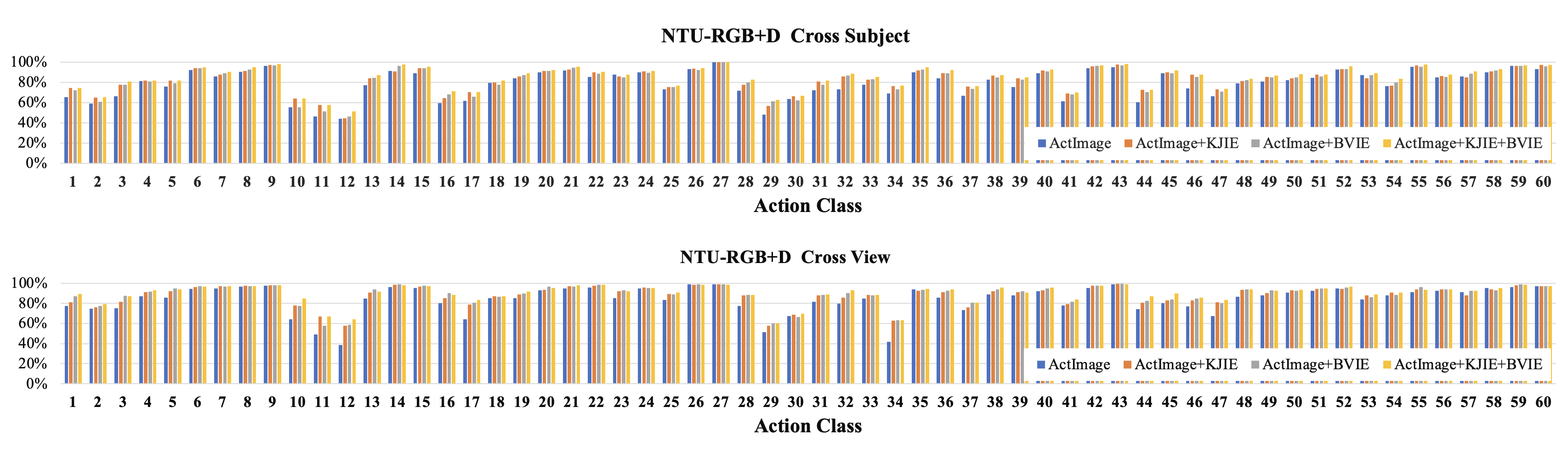}
\caption{Class by class action recognition accuracy comparisons of AF-CNN with and without KJFE, BVFE by using cross-subject and cross-view metrics on the NTU RGB+D dataset}
\label{fig:actionaccuracy}
\end{figure}

From the comparison of class by class action recognition accuracy as shown in Fig~\ref{fig:actionaccuracy}, we can find that our both KJFE and BVFE improve the action recognition accuracy on each action class. 
 It is worth mentioning that on challenging action classes (e.g., class 11 reading and class 12 writing), our KJFE+BVFE improve the accuracy by 36.2\% and 66.8\% respectively on cross-view metric. 
For other action classes, our KJFE+BVFE remains high action recognition accuracy.

\subsubsection{Ablation study on joint velocity transform module}

\begin{table}[h]
\small
\centering
\caption{Ablation study on joint velocity transform module. The AFE-CNN only uses key joints feature enhanced images as inputs, JVTM denotes the joint velocity transform module.}
\resizebox{80mm}{!}{
\begin{tabular}{l|c|cc|l}
\toprule
\multicolumn{1}{c|}{\multirow{2}{*}{\textbf{Methods}}} &
\multicolumn{2}{c}{\textbf{Accuracy (\%)}} \\ \cline{2-3}
\multicolumn{1}{c|}{} &
\multicolumn{1}{c}{\textbf{CrossSubject}} &
\multicolumn{1}{c}{\textbf{CrossView}}  \\ \midrule
AFE-CNN  & 78.9 & 85.6  \\
AFE-CNN+JVTM   & \textbf{86.2} & \textbf{92.2} \\ \bottomrule
\end{tabular}}
\label{ablationstudyonjvtm}
\end{table}

Here we design an ablation study on the joint velocity transform module to analysis its contribution. 
As shown in Table~\ref{ablationstudyonjvtm}, there is a large gap between whether the AFE-CNN uses the joint velocity transform module. 
For example, the JVTM improves  the accuracy on cross-subject and cross view evaluation metrics by 7.3\% and 6.6\%, respectively. 
Thus, we believe that the joint velocity transform module plays a key role in achieving outstanding results in performing CNN-based action recognition task as it provides motion information which is critical to recognize actions.

\subsection{Complexity measurement}
In this section, we analysis the complexity of our AFE-CNN by measuring its computing time, floating-point operations per second (Flops) and further compare with several representative methods. 
It is noted that a comparison of processing time cannot be done fairly due to the diversity in use of frameworks (e.g., TensorFlow, PyTorch) and computing platforms (e.g., single GPU and multi-GPUs) are very diverse.

\begin{table}[h]
\small
\centering
\caption{Methods comparison of runtime and resource consumption. Flops denotes the floating-point operations per second.}
\resizebox{100mm}{!}{
\begin{tabular}{l|c|c|c|cc|l}
\toprule
\multicolumn{1}{c|}{\textbf{Methods}} &
\multicolumn{1}{c|}{\textbf{Architecture}} &
\multicolumn{1}{c|}{\textbf{GPUs}} &
\multicolumn{1}{c|}{\textbf{Times (ms)}} &
\multicolumn{1}{c}{\textbf{Flops}} \\  \midrule
Synthesized CNN~\cite{SynthesizedCNN} & CNN & GTX 1080 GPU & $\sim$ 390ms & - \\
SPMF+ResNet~\cite{SPMFInception-ResNet-222} & CNN & GTX 1080Ti GPU & $\sim$ 128ms & 13.0GFlops \\
ST-GCN~\cite{stgcn} & GCN & Tesla K80 GPU & $\sim$ 93ms & 16.3GFlops\\
PoT2I+Inception-v3~\cite{PoT2I+Inception-v3} & CNN & 2 × GTX 1080Ti GPU  & $\sim$ 38ms & 6.0GFlops\\ \midrule 
AFE-CNN  & CNN & RTX 3090 GPU & $\sim$ 3.5ms & 1.4GFlops\\ \bottomrule
\end{tabular}}
\label{computetimecost}
\end{table}

As shown in Table~\ref{computetimecost}, our AFE-CNN only cost 3.5ms for one forward inference and only consumes 1.4GFlops, which are significantly lower than other methods.  
Compared with other methods~\cite{PoT2I+Inception-v3,SPMFInception-ResNet-222,SynthesizedCNN} using handcrafted action feature images, 
our AFE-CNN can be fully executed on GPU. Since we adopt a light-weight CNN model as the back bone, our method can minimize the computing time. Although the GCN-based model has applaudable performance on 3D skeleton-based action recognition task, it costs more computing time due to its complex computational structure.

\subsection{Visualization of Action Feature Images}
To illustrate our action feature enhance mechanism, we visualize the action feature enhanced images and multi-frame attention maps of drink water and jump up actions. We further compare them with the action feature images without utilizing any proposed feature enhance module. 

\begin{figure}[h]
\centering
\includegraphics[height=4.5cm ,width=12cm,angle=0]{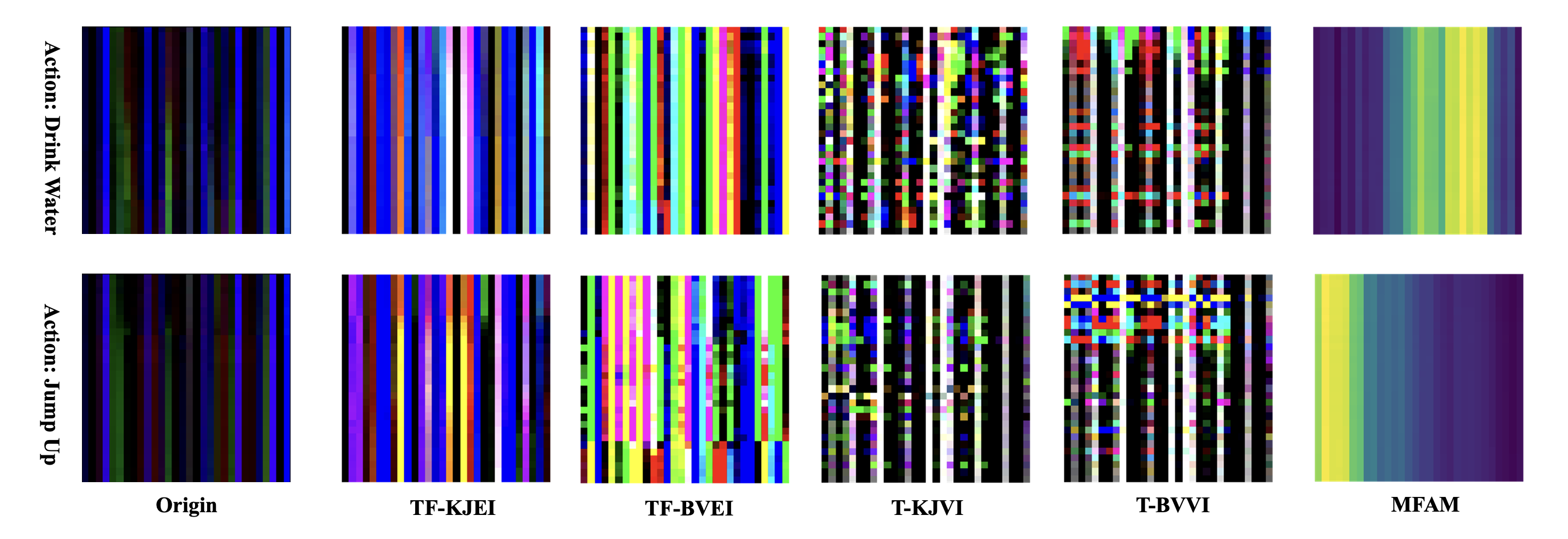}
\caption{Visualization of action feature images, the origin denotes the images generated without any proposed action feature enhance module.}
\label{fig:actionimages}
\end{figure}

As shown in Fig~\ref{fig:actionimages}, we can observe that the strips in action feature enhanced images are much more distinguishable than the images without being enhanced by our proposed feature enhance modules.
Moreover, it is obviously that the TF-KJEI, TF-BVEI, T-KJVI, and T-BVVI encode different information in an action. 
Thus the CNN model can lead to outstanding performance thanks to the rich features in these action feature enhanced images. 
For MFAM, we can find that it successfully finds the key frames in an action sequence, where the yellow color indicates a stronger attention.

\section{Conclusion}
In this paper, we have proposed a novel learning-based action feature enhanced method for 3D skeleton-based action recognition, which namely AFE-CNN.
Firstly, our AFE-CNN enhances the action features from key joint and bone vector perspectives to adapt to various camera views and body sizes.
Secondly, the key frames of a skeleton sequence is enhanced by devicing a multi-frame attention module and a temporal embedding module to enhance temporal information. 
Thanks to the action feature enhance modules, our AFE-CNN effectively overcome the limitations of handcraft action features. 
The extensive experimental results demonstrate that our AFE-CNN achieves state-of-the-art performance on three benchmark datasets, and the recognition accuracy of challenging action classes is significantly improved. 
Notably, our AFE-CNN adopts light-weight CNN models so that the required computational load and computing time are extremely low. 
This makes it a good candidate technique for real-world applications.

\bibliography{mybibfile}

\end{document}